\documentclass[10pt,twocolumn,letterpaper]{article}

\usepackage[pagenumbers]{cvpr} 

\usepackage[dvipsnames]{xcolor}

\usepackage{amssymb}
\newcommand{\cmark}{\checkmark}  
 \usepackage{multirow}
 \usepackage[numbers]{natbib}

\definecolor{cvprblue}{rgb}{0.21,0.49,0.74}
\usepackage[pagebackref,breaklinks,colorlinks,citecolor=cvprblue]{hyperref}

\def\paperID{*****} 
\def\confName{CVPR}
\def\confYear{2024}

\title{Evolution of ReID: From Early Methods to LLM Integration}

\author{
Amran Bhuiyan$^{1}$\thanks{Corresponding authors: \texttt{\{amran,jhuang\}@yorku.ca}} \quad
Mizanur Rahman$^{1,2}$ \quad
Md Tahmid Rahman Laskar$^{1,2}$ \\
Aijun An$^{2}$ \quad
Jimmy Xiangji Huang$^{1}$ \textsuperscript{*} \\
$^{1}$Information Retrieval and Knowledge Management Research Lab, York University, Toronto, Canada \\
$^{2}$Department of Electrical Engineering and Computer Science, York University, Toronto, Canada
}

\begin{document}
\maketitle
\begin{abstract}
Person re-identification (ReID) has evolved from handcrafted feature-based methods to deep learning approaches and, more recently, to models incorporating large language models (LLMs). Early methods struggled with variations in lighting, pose, and viewpoint, but deep learning addressed these issues by learning robust visual features. Building on this, LLMs now enable ReID systems to integrate semantic and contextual information through natural language. This survey traces that full evolution and offers one of the first comprehensive reviews of ReID approaches that leverage LLMs, where textual descriptions are used as privileged information to improve visual matching. A key contribution is the use of dynamic, identity-specific prompts generated by GPT-4o, which enhance the alignment between images and text in vision-language ReID systems. Experimental results show that these descriptions improve accuracy, especially in complex or ambiguous cases. To support further research, we release a large set of GPT-4o-generated descriptions for standard ReID datasets. By bridging computer vision and natural language processing, this survey offers a unified perspective on the field's development and outlines key future directions such as better prompt design, cross-modal transfer learning, and real-world adaptability
\end{abstract}    
\section{Introduction}

The field of artificial intelligence (AI) is marked by its rapid evolution, continually adapting to meet the evolving needs of modern society. Among the broad spectrum of AI technologies, person ReID emerges as a crucial technology, significantly enhancing security and surveillance systems. At its core, ReID involves consistently associating an individual's identity across a network of spatially disparate, non-overlapping camera feeds, 
which poses substantial challenges due to its inherent complexity. This complexity arises from several intrinsic and extrinsic factors, including variability in human body postures, diversity in camera viewpoints, fluctuations in ambient lighting conditions, and the existence of perceptual gaps across different data modalities. These factors contribute to substantial visual discrepancies in representing the same individual across different instances. Conversely, distinct individuals may present remarkably similar visual appearances, further complicating the ReID process.

\begin{figure*}[t!]
\centering
\includegraphics[width=0.9\linewidth,trim={0in 0in 0in 0in},clip]{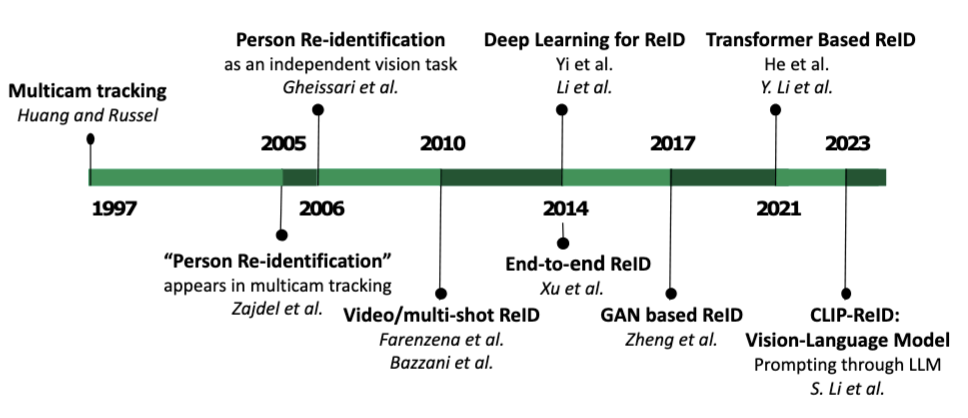}

   \caption{Milestones in the person ReID history.}
\label{fig:mile}
\vspace{-4mm}
\end{figure*}

ReID has undergone a significant transformation since its origins in multicamera tracking systems~\cite{Wang2013}, evolving into a distinct and mature area within computer vision. The conceptual foundation was laid in 1997 when Huang and Russell~\cite{huang97} introduced Bayesian frameworks to model appearance transitions across non-overlapping camera views. This initial effort was extended in 2005 by Zajdel et al.~\cite{Zaj2005}, who formalized the task of explicitly recognizing individuals across different camera perspectives. In 2006, Gheissari et al.~\cite{Gheissari2006} advanced the field through image-based ReID using visual features and segmentation techniques. The focus then shifted towards multi-shot and video-based methods around 2010, as exemplified by Bazzani et al.~\cite{bazzani2010} and Farenzena et al.~\cite{Farenzena}, who leveraged temporal information and frame-level diversity to enhance matching performance. A paradigm shift occurred with the introduction of deep learning, particularly following the work of Krizhevsky et al.~\cite{Krizhevsky} in 2012, which led to the adoption of convolutional neural networks in ReID. This development was rapidly embraced in 2014 through the application of Siamese networks~\cite{DYi2014, wli2014}, enabling more effective feature learning for pairwise identity comparison, while Xu et al.~\cite{yxu2014} introduced end-to-end ReID systems that unified detection and recognition tasks. Subsequent advances included the use of Generative Adversarial Networks (GANs) for data augmentation and domain adaptation~\cite{Zheng2017}, and the integration of Transformers to model spatial and sequential dependencies~\cite{she2021}. Most recently, a new direction has emerged through the incorporation of vision-language models, with Li et al.~\cite{lsi2023} demonstrating how aligning textual and visual modalities can significantly enhance identity representation and retrieval. This transition toward multimodal ReID marks a convergence between computer vision and natural language processing, leading to a new generation of context-aware and interpretable systems. The trajectory of this progression, from early probabilistic models and hand-crafted features to deep learning, generative approaches, and integration of vision languages, is summarized in Figure~\ref{fig:mile} and builds on foundational surveys~\cite{zheng2016person, Leng2020} while expanding the scope to recent innovations.

\subsection{Scope and Objectives}

ReID has seen a rapid expansion in both methodological innovation and application domains, yet existing surveys predominantly focus on visual-only techniques, such as handcrafted descriptors~\cite{vaz_acm2013, bedagkar2014survey}, CNN-based pipelines~\cite{wu2019deep, wang2018survey}, and modality-specific advancements like thermal~\cite{huang2023deep}, gait~\cite{athira_acm2019}, or video ReID~\cite{saad2024deep}. While these studies have substantially shaped the field, they operate within a unimodal paradigm that overlooks recent shifts toward integrating semantic-level information through language. With the growing emergence of large language models (LLMs) and vision-language models~\cite{zhu2023minigpt, bai2023qwen, liu2024visual}, the landscape of person ReID is undergoing a transformative evolution. Textual prompts and LLM-generated descriptions now offer an enriched modality that supports contextual identity matching, label-efficient training, and human-in-the-loop feedback, yet these innovations remain underexplored in prior literature.

\begin{table*}[t!]
\centering
\caption{Major person-ReID surveys (2013–2025).}
\label{tab:survey_overview}
\renewcommand{\arraystretch}{0.8}
\resizebox{\textwidth}{!}{%
\begin{tabular}{|p{5.5 cm}|p{6.2cm}|p{4cm}|c|}
\hline
\textbf{Survey (Year, Venue)} & \textbf{Main Focus} & \textbf{Re-ID Settings Covered} & \textbf{LLM} \\ \hline
Vezzani et al.\ 2013, ACM CSUR \cite{vaz_acm2013} & Hand-crafted features; distance metrics (pre-DL) & $\cmark{}$ RGB single-shot & No \\ \hline
Saghafi et al.\ 2014, IET CV \cite{saghafi2014review} & Holistic vs.\ part-based appearance; metric learning & $\cmark{}$ RGB single-/multi-shot & No \\ \hline
Bedagkar-Gala \& Shah 2014, IVC \cite{bedagkar2014survey} & Colour/texture descriptors; metric fusion & $\cmark{}$ RGB closed-world & No \\ \hline
Wang et al.\ 2018, CAAI TIT \cite{wang2018survey} & CNN + GAN pipelines & $\cmark{}$ RGB single-shot & No \\ \hline
Almasawa et al.\ 2019, IEEE Access \cite{almasawa2019survey} & Deep CNN systems; image/video hybrids & $\cmark{}$ RGB single-shot, $\cmark{}$ Video & No \\ \hline
Wu et al.\ 2019, Neurocomputing \cite{wu2019deep} & Six DL branches (ID, verification, metric, component, video, augmentation) & $\cmark{}$ RGB single-shot & No \\ \hline
Nambiar et al.\ 2019, ACM CSUR \cite{athira_acm2019} & Gait biometrics & $\cmark{}$ Gait / video sequences & No \\ \hline
Leng et al.\ 2020, TCSVT \cite{Leng2020} & Open-set / unconstrained ReID & $\cmark{}$ RGB single-shot, $\cmark{}$ Partial/occluded & No \\ \hline
Mathur et al.\ 2020, ICETCE \cite{mathur2020brief} & View / lighting / low-resolution challenges & $\cmark{}$ RGB single-shot & No \\ \hline
Islam 2020, IVC \cite{islam2020person} & Deep-metric losses & $\cmark{}$ RGB single-shot & No \\ \hline
Wang et al.\ 2021, AIC \cite{wang2021cross} & Cross-Domain ReID & $\cmark{}$ RGB single-shot & No \\ \hline
Yang et al.\ 2021, CDS \cite{yang2021survey} & Unsupervised Domain Adaptation (UDA) & $\cmark{}$ RGB single-shot & No \\ \hline
Ye et al.\ 2021, TPAMI \cite{deep_2022} & Closed-vs-open CNN ReID; AGW baseline & $\cmark{}$ RGB single-shot & No \\ \hline
Meng et al.\ 2022, IVC \cite{ming2022deep} & Four DL families (GAN, adv., metric, local) & $\cmark{}$ RGB single-shot & No \\ \hline
Huang et al.\ 2022, Inf.\ Fusion \cite{huang2023deep} & Cross-modality ReID & $\cmark{}$ RGB-IR single-shot & No \\ \hline
Li et al.\ 2022, IEEE TBIS \cite{li2022overview} & Deep Reinforcement Learning-based ReID & $\cmark{}$ RGB single-shot & No \\ \hline
Singh et al.\ 2022, MTAP \cite{singh2022comprehensive} & Multiple aspects & $\cmark{}$ RGB, $\cmark{}$ Video, $\cmark{}$ RGB-IR & No \\ \hline
Liu et al.\ 2023, Mathematics \cite{liu2023survey} & Cross-modal ReID (math view) & $\cmark{}$ RGB-IR, $\cmark{}$ Thermal, $\cmark{}$ Text-image, $\cmark{}$ Sketch-photo & No \\ \hline
Peng et al.\ 2024, ACM TOMM \cite{peng2023deep} & Occluded / partial ReID & $\cmark{}$ Partial / occluded & No \\ \hline
Saad et al.\ 2024, EURASIP JASP \cite{saad2024deep} & Video-based ReID & $\cmark{}$ Video ReID & No \\ \hline
Qian et al.\ 2024, SN Comp Sci \cite{qian2024identifying} & Person + vehicle ReID; transformers & $\cmark{}$ RGB closed/open, $\cmark{}$ Vehicle ReID & No \\ \hline
Sarker et al.\ 2024, IEEE T-IV \cite{sarker2024} & Vision-Transformer ReID & $\cmark{}$ RGB single-shot & No \\ \hline
Chang et al.\ 2024, MTAP \cite{chang2024comprehensive} & Visible–Infrared ReID & $\cmark{}$ RGB $\leftrightarrow$ IR / thermal
 & No \\ \hline
Zahra et al.\ 2023, Pattern Recog. \cite{zahra2023person} & Supervised / unsup. / semi-sup. ReID & $\cmark{}$ RGB single-shot & No \\ \hline
Asperti et al.\ 2025, MVA \cite{asperti2025review} & Supervised vs.\ unsupervised recent ReID & $\cmark{}$ RGB single-shot, $\cmark{}$ Unsupervised / UDA & No \\ \hline
\textbf{This Survey} & Full history; first text-prompt / LLM survey & $\cmark{}$ RGB single-shot (text-prompt) & \textbf{Yes} \\ \hline
\end{tabular}}
\end{table*}

This survey aims to bridge that gap by providing the first comprehensive review that situates prompt-driven and LLM-guided ReID within the broader historical trajectory of the field. We trace developments from early RGB-based matching~\cite{huang97, Gheissari2006} to the current paradigm of multimodal, language-informed models~\cite{lsi2023}. Unlike previous works that focus on a single axis, such as model architecture, data modality, or training strategy, our survey synthesizes these advancements under the emerging vision-language umbrella. Specifically, we highlight how LLMs inject semantic understanding into identity representation, feasibility of zero-shot generalization, and enable dynamic textual feedback mechanisms. We also identify open research challenges, including prompt robustness, domain-adaptive multimodal learning, and cross-modal distillation, which we believe are critical for the next phase of ReID research. By consolidating these trends, our work not only extends the scope of earlier surveys~\cite{zheng2016person, Leng2020} but also proposes a unified, forward-looking framework to support future innovations in language-aware ReID systems.

In addition to revisiting key milestones in the evolution of ReID, this survey introduces a new perspective grounded in language-driven techniques. Specifically, we explore how LLMs can be used to generate extended, descriptive prompts that provide richer contextual information for identity representation. These LLM-generated prompts go beyond simple textual labels by offering dynamic, image-specific semantic cues that strengthen the connection between visual and language modalities. When integrated into models such as CLIP-ReID~\cite{lsi2023}, these prompts lead to moderate improvements in distinguishing visually similar individuals, particularly in challenging or low-resolution settings. While the performance gains are incremental, the approach highlights a novel and promising direction for enhancing semantic alignment in ReID tasks. To support this line of research, we also generate and publicly release a collection of detailed natural language descriptions for the widely used Market1501~\cite{zheng_mar2015} and DukeMTMC-reID~\cite{Ristani2016} datasets using the GPT-4o model. These prompts introduce a new semantic layer for training and evaluation, providing the community with a standardized and accessible benchmark for exploring prompt-based and multimodal ReID systems.

\subsection{Contributions}

This survey introduces a new direction in the field of person ReID by exploring how LLMs can enhance identity recognition through text-based semantic guidance. Our main contributions are outlined below:

\begin{itemize}
    \item  This is the first survey to focus specifically on the role of LLMs in person ReID. We explore how LLMs, through prompt generation and semantic understanding, can improve identity matching by providing rich textual context beyond traditional visual features.

    \item We connect earlier ReID approaches - from handcrafted features and CNN-based architectures to GANs and Transformers- with the latest advancements that use language prompts and vision-language alignment. This creates a unified view of the field's progression and helps readers understand how LLMs are shaping the next phase of ReID research.

\item To encourage further research, we generate and publicly release descriptive prompts\footnote{\url{https://github.com/mdamranhossenbhuiyan/Extended_Prompt_CLIP_REID}} created using GPT-4o for two widely used ReID datasets—Market1501~\cite{zheng_mar2015} and DukeMTMC-reID~\cite{Ristani2016}. These descriptions introduce a new semantic layer, making evaluating and developing LLM-guided ReID systems easier.

\end{itemize}

\subsection{Comparison with Existing Surveys}

Over the past decade, numerous surveys have played an important role in shaping the landscape of person ReID research. Early contributions, such as those by Vezzani et al.~\cite{vaz_acm2013}, Saghafi et al.~\cite{saghafi2014review}, and Bedagkar-Gala and Shah~\cite{bedagkar2014survey}, primarily focused on handcrafted features and metric learning methods in the pre-deep learning era. As the field evolved with the introduction of deep neural networks, surveys began to explore various deep learning paradigms, ranging from CNN and GAN based architectures~\cite{wang2018survey, almasawa2019survey}, to taxonomy driven reviews of six major deep ReID branches~\cite{wu2019deep}, and explorations of deep metric losses~\cite{islam2020person}. Other works addressed domain specific challenges such as video based ReID~\cite{saad2024deep}, gait analysis~\cite{athira_acm2019}, and occlusion robustness~\cite{Leng2020, peng2023deep}. Cross-modal surveys further expanded the scope to infrared and thermal domains~\cite{huang2023deep, chang2024comprehensive, liu2023survey}, while others reviewed ReID under open set settings~\cite{Leng2020}, person and vehicle scenarios~\cite{qian2024identifying}, deep reinforcement learning~\cite{li2022overview} and Transformer-based models~\cite{sarker2024}. Despite this breadth, each prior survey generally advanced the field along a single direction, whether by data modality, learning strategy, or application constraint, without integrating broader multimodal perspectives.

What distinctly sets this survey apart is its introduction of the language-driven paradigm, which has not been addressed in any prior work to date. As summarized in Table~\ref{tab:survey_overview}, all existing surveys, spanning from 2013 to 2025, are based exclusively on visual modalities and do not consider the role of text or language in identity representation. None of the previous works incorporates LLMs, textual prompts, or the emerging capabilities of vision and language fusion, despite their growing relevance in other AI domains. In contrast, this survey not only provides the first complete historical overview from early RGB-based single-shot methods to prompt-driven retrieval, but also systematically explores how LLMs enrich ReID by enhancing semantic understanding, supporting zero-shot generalization, and enabling natural language interaction in human-centered systems. Furthermore, by contextualizing these advances within a broader taxonomy and projecting key future directions, such as prompt design, cross modal distillation, and domain adaptive multimodal learning, this work establishes a new trajectory for ReID research and offers a consolidated foundation for the next generation of language aware and semantically grounded ReID systems.


\section{Problem Definition of ReID}
\label{prob_def}
ReID is a recognition or matching task that identifies a query image \( q \) from a gallery set \( G = \{g_1, g_2, \dots, g_n\} \). The goal is to find the best match \( g^* \) by minimizing the distance \( d(\mathbf{f}(q), \mathbf{f}(g_i)) \) between their feature representations:
\begin{equation}
g^* = \arg\min_{g_i \in G} d(\mathbf{f}(q), \mathbf{f}(g_i))
\label{eqn:reid}
\end{equation}

Here, \( d(\cdot) \) represents a distance metric like Euclidean distance or cosine similarity. ReID methods differ in how they extract the feature embedding \( \mathbf{f} \). Traditional approaches use handcrafted features, while deep learning-based methods leverage neural networks for representation learning. The system outputs a ranked list of gallery images, ordered by their similarity to the query. Fig~\ref{fig:tax} summarizes different ReID approaches based on their feature extraction techniques.

\section{Survey Methodology}
\begin{figure*}[t!]
\centering
\includegraphics[width=1.1\linewidth, trim=2.8cm 0cm 0cm 0cm, clip]{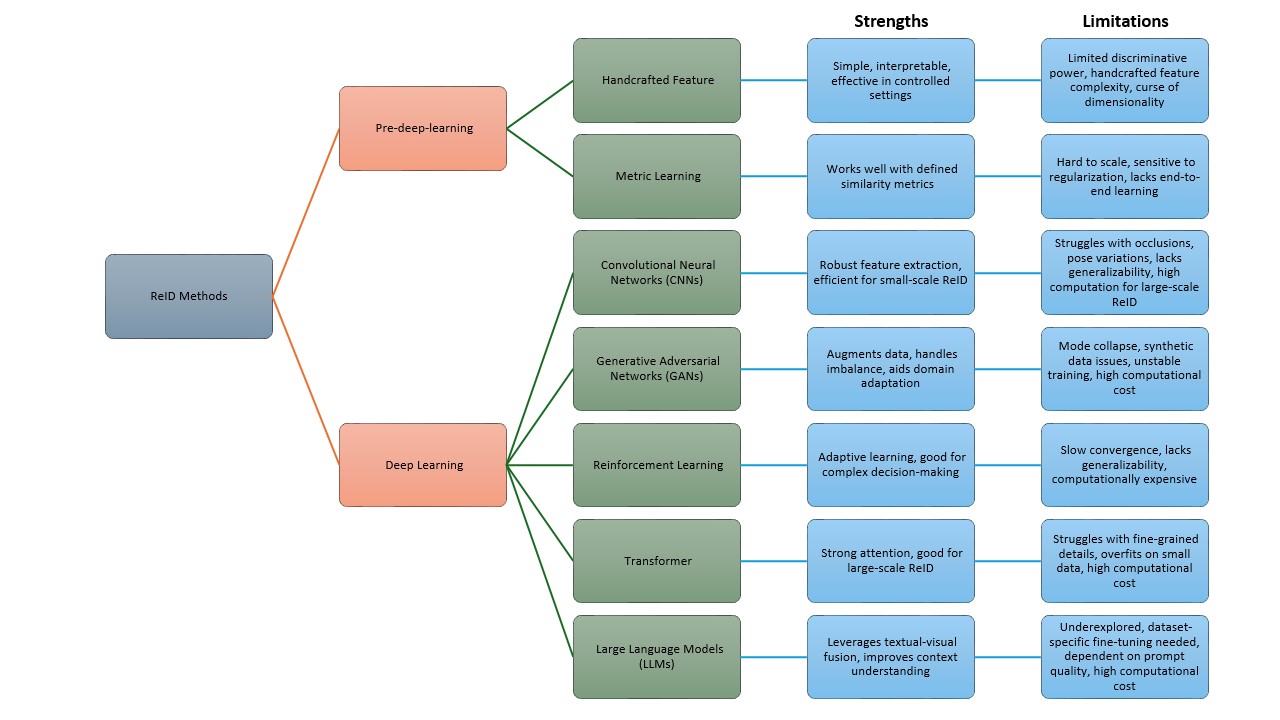}

   \caption{A taxonomy of person ReID methods categorized into pre-deep learning and deep learning approaches. The taxonomy further distinguishes between specific techniques such as handcrafted features, metric learning, CNNs, GANs, reinforcement learning, transformers, and LLMs. For each category, corresponding strengths and limitations are also summarized, offering a comparative perspective on the evolution of ReID methodologies.}
\label{fig:tax}

\end{figure*}

In this paper, we explore the dynamic and evolving field of person ReID technologies through a comprehensive and methodologically rigorous survey. Unlike earlier works that focus on narrow aspects such as handcrafted descriptors~\cite{TG_sur2012, AS_sur2014, SG_2014}, traditional matching pipelines~\cite{Satta2013}, or specific deep learning techniques~\cite{zheng2016person, karanam_pami, deep_2022, sarker2024}, our survey presents a whole view of the ReID landscape. This includes a historical transition from pre-deep learning paradigms to modern deep architectures, such as convolutional neural networks, transformers, and more recently, LLMs. As illustrated in Figure~\ref{fig:tax}, we present a structured taxonomy that categorizes ReID methods based on learning strategies and outlines their respective strengths and limitations. Our survey is also one of the first to explore the emerging role of LLMs in ReID systems, investigating their potential for bridging vision-language gaps, enabling prompt-driven identification, and supporting zero-shot or human-in-the-loop applications. To that end, we conduct experimental validation using GPT-4o~\cite{gpt4_technicalreport} to assess its relevance and applicability in ReID tasks.

To ensure objectivity and transparency in the selection of literature, we adopt the PRISMA (Preferred Reporting Items for Systematic Reviews and Meta-Analyses) protocol~\cite{shamseer2015preferred}, which provides a structured and reproducible guideline for identifying and screening relevant studies. We restricted our review to top-tier journal and conference publications that report quantitative results on widely recognized benchmark datasets-namely VIPeR~\cite{gray2008}, Market1501~\cite{zheng_mar2015}, and DukeMTMC-reID~\cite{Ristani2016}. These datasets are not only extensively used by the research community but also facilitate fair and consistent comparison of ReID models. Articles that lacked standard evaluation metrics such as mean Average Precision (mAP) and Cumulative Matching Characteristic (CMC) Rank-1, or relied on lesser-known datasets with limited community adoption, were excluded. This rigorous selection process enables us to synthesize performance trends, model limitations, and research gaps with a high degree of confidence, ultimately contributing a reliable and forward-looking analysis of the ReID domain (see Table~\ref{tab:trad_reid} \& \ref{tab:deep} for a comparative breakdown).

\subsection{Pre-Deep Learning Approaches to ReID}

Person ReID systems fundamentally rely on two core components: a robust feature embedding function \textbf{f} and an effective distance metric $d$, as outlined in Equation~\ref{eqn:reid}. Before the deep learning era, most ReID pipelines were designed by decoupling these components into modular stages: (i) crafting discriminative visual representations, and (ii) learning or applying a distance function to compare these representations. This section revisits the two dominant technical families from this pre-deep era - \textit{handcrafted feature-based representation learning} and \textit{metric learning}- along with their subcategories and performance profiles.

\subsubsection{Handcrafted Representation Learning}
Handcrafted approaches primarily focus on human-designed features that encode the spatial, chromatic, and textural characteristics of pedestrians. Global descriptors capture entire pedestrian images using color histograms and edge or gradient patterns. For instance, ELF~\cite{Gheissari2006} used spatial-temporal segmentation and color-edge histograms to construct holistic signatures. Similarly, the work by~\cite{wang2007} incorporated color co-occurrence matrices for body regions, which were effective under viewpoint constraints. Recognizing limitations of global methods,  a significant milestone in ReID research was achieved with the introduction of the \textit{Symmetry-Driven Accumulation of Local Features (SDALF)} framework~\cite{Farenzena}, based on the idea that features closer to the body symmetry axes are more robust against scene clutter and body extremities. Similar features, such as \textit{MSCR (Maximally Stable Color Regions)} and color histograms, are employed in prior works~\cite{cheng2011,bhuiyan2014,bhuiyan2015,buiyan2015_b,mirmahaboub2015,bhuiyan2018exploiting}, where body parts are segmented using \textit{Custom Pictorial Structures (CPS)}~\cite{cheng2011} and \textit{Stel Component Analysis (SCA)}~\cite{bhuiyan2014}, while appearance descriptors are extracted from these segmented regions for matching signatures. Later, the \textit{Local Maximal Occurrence (LOMO)} descriptor \cite{Liao2015} introduced scale-invariant local features and max pooling techniques to handle illumination variations, further solidifying its effectiveness in ReID benchmarks. The \textit{Gaussian of Gaussian (GoG)} feature \cite{matsukawa} extended this idea by modeling color and texture cues through hierarchical Gaussian distributions, leading to more discriminative feature representations, but at the cost of extremely high feature dimensions (26k-27k). Attribute-based representations attempted to bridge the semantic gap by annotating pedestrians with human-interpretable traits (e.g., clothing type, gender). Layne et al.~\cite{layne2012} manually labeled binary attributes, while Su et al.~\cite{csu2015} embedded them into continuous low-rank spaces. These methods showed better generalization across datasets but required extensive labeling or external data sources.

\paragraph{Strengths and Limitations} Handcrafted features offer simplicity and interpretability, and require no large-scale training. However, they suffer from limited expressiveness, poor scalability, and reliance on high-dimensional encodings. Their performance is sensitive to illumination, viewpoint, and background clutter, making them unsuitable for complex or real-time ReID applications.

\begin{table*}[t!] 
    \centering
    \caption{Comparative analysis of traditional ReID approaches on VIPeR and Market1501 datasets.}
    
    \resizebox{0.9\textwidth}{!}{
    
  \begin{tabular}{c|c|c|c}

    \hline
  {\textbf{Pre-deep-learning Methods}} & \textbf{References} & \textbf{VIPeR (Rank-1)} & \textbf{Market1501 (Rank-1)} \\ \hline

    ELF~\cite{Gheissari2006} & CVPR,06 & 12.00 & - \\
    KISSME~\cite{kostin2012} & CVPR,12 & 19.60 & 44.42 \\
    SDALF~\cite{Farenzena} & CVPR,10 & 19.87 & 20.50 \\
    SCA+MSCR~\cite{bhuiyan2014} & ECCV,14 & 21.67 & -- \\
    SCNCD~\cite{Yang2014} & ECCV,14 & 37.80 & - \\
    LOMO~\cite{Liao2015} & CVPR,15 & 40.00 & 43.23 \\
    MLAPG~\cite{liao_psd2015} & ICCV,15 & 40.73 & 42.52 \\
    EquiDML~\cite{wang_pr2017} & PR,17 & 40.93 & 53.36 \\
    NullSpace~\cite{zhang2016} & CVPR,16 & 42.28 & 54.60 \\
    SalMat~\cite{zhao2013-1} & TPAMI,17 & 44.56 & - \\
    GoG~\cite{matsukawa} & CVPR,16 & \textbf{49.70} & 59.77 \\ \hline

   {\textbf{Early Deep Learning-Based Methods}} & \textbf{References} & \textbf{VIPeR (Rank-1)} & \textbf{Market1501 (Rank-1)} \\ \hline
    DML~\cite{DYi2014} & ICPR,14 & 28.20 & - \\
    End-to-End (AlexNet)~\cite{yxu2014} & Arxiv,14 & 41.50 & 55.10 \\
    End-to-End (VGG-16)~\cite{yxu2014} & Arxiv,14 & 47.50 & 60.30 \\
    ConvNet~\cite{Ahmed} & CVPR,15 & 34.80 & - \\
    SI-CI~\cite{wang_cvpr2016} & CVPR,16 & 35.80 & - \\
    MCP-CNN~\cite{Cheng_cvpr2016} & CVPR,16 & 47.80 & 45.10 \\ 
    Gated-Siamese~\cite{varior_eccv2016} & ECCV,16 & 37.80 & \textbf{65.88} \\ \hline

  \end{tabular}
  }
  \label{tab:trad_reid}
  \vspace{-2mm}
\end{table*}

\subsubsection{Distance Metric Learning}
Given a fixed feature space, metric learning methods aim to optimize the similarity function $d$ that brings features of the same identity closer and pushes others apart. These methods often optimize a Mahalanobis distance through various strategies.  Early work by~\cite{epxing2003} approached this through a convex optimization framework, which was later simplified in \textit{Keep It Simple and Straightforward MEtric (KISSME)} ~\cite{kostin2012}, which bypasses subspace learning via a likelihood ratio test. ~\cite{zhu2017} improved its robustness by leveraging \textit{Gaussian Mixture Models (GMMs)}. Other techniques relaxed positive semidefinite constraints for computational efficiency~\cite{hirzer2012} or introduced bilinear similarities to capture cross-patch relations~\cite{chen2015}. \cite{liao_psd2015} further refined the process by assigning different weights to positive and negative samples under positive semidefinite constraints, while ~\cite{wang_pr2017} developed \textit{EquiDistance constrained Metric Learning (EquiDML)} to map images of the same person to a single vertex in a simplex. Subspace-based methods like \textit{Cross-view Quadratic Discriminant Analysis (XQDA)}~\cite{Liao2015} and \textit{Null Space Discriminant Analysis(NullSpace)}~\cite{zhang2016} project features into discriminative subspaces before applying metric learning, reducing intra-class variance and enhancing inter-class separability. These ongoing advancements improve both the scalability and generalization of ReID models by refining the embedding space for more robust identity discrimination.

In addition to traditional metric learning, \textit{Support Vector Machines (SVMs)} and boosting have also been explored for person ReID. ~\cite{prosser2010} combined multiple weak RankSVMs to form a stronger ranking model, while ~\cite{zhang_svm2016} trained a separate SVM for each identity, improving the mapping from test images to their feature space. AdaBoost-based boosting approach was also used to merge various simple features into a more robust similarity function~\cite{gray2008}.

\paragraph{Strengths and Limitations}
While comparing with handcrafted-based ReID approaches, metric learning methods were effective in scenarios with moderate data availability and carefully designed features. However, they required extensive pairwise labeling and could not jointly optimize feature extraction, leading to suboptimal performance when feature quality was poor. They were also computationally expensive for large-scale deployments.

\subsubsection{Quantitative Comparison and Analysis}
Table~\ref{tab:trad_reid} compares traditional and early deep-learning ReID methods on VIPeR~\cite{gray2008} and Market1501~\cite{zheng_mar2015}, illustrating the gradual improvements in accuracy and the shift toward data-driven models. Among traditional approaches, ELF~\cite{Gheissari2006} was an early baseline with a Rank-1 of 12.0\% on VIPeR, which steadily grew with methods like \textit{KISSME}~\cite{kostin2012}, \textit{SDALF}~\cite{Farenzena}, and \textit{LOMO}~\cite{Liao2015}. Ultimately, \textit{GoG}~\cite{matsukawa} achieved the highest VIPeR Rank-1 of 49.70\%, also outperforming others on Market1501 with 59.77\%. In parallel, early deep-learning techniques struggled on VIPeR due to limited training data, with \textit{Deep Metric Learning (DML)}~\cite{DYi2014} only reaching 28.20\% Rank-1 and \textit{ConvNet}~\cite{Ahmed} at 34.80\%. However, once larger datasets were introduced, performance improved substantially; End-to-End \textit{(VGG-16)}~\cite{yxu2014} attained 60.30\% on Market1501, and \textit{Gated-Siamese}~\cite{varior_eccv2016} surpassed previous models with 65.88\%. These trends highlight how early deep-learning methods began to rival or exceed handcrafted techniques on more extensive benchmarks, opening the way for further advancements in representation learning and model scalability.

This comparison illustrates how pre-deep-learning ReID methods relied heavily on engineered representations and metric tuning, while early deep networks gradually improved with better data availability. The transition also underscores the limitations of modular design in traditional pipelines, especially under domain shifts. These insights provide a foundation to understand why unified learning frameworks became necessary for further progress.

\subsubsection{Implications and Transition to Deep Learning}
The performance plateau of  methods, especially under challenging real-world scenarios, underscored several limitations: rigid handcrafting, modular pipeline decoupling, lack of semantic abstraction, and inefficient generalization. These factors catalyzed the transition toward deep learning, where convolutional architectures jointly optimize feature and similarity functions in an end-to-end manner. In the next section, we elaborate on this shift and explore how deep networks addressed the above challenges to dominate modern ReID systems.

\subsection{Deep Learning-Based Approaches}
\label{deep}

The field of person ReID has grown rapidly, especially after the introduction of deep convolutional networks in visual recognition~\cite{Krizhevsky}. Deep learning helped overcome the limitations of hand-crafted features by allowing models to automatically learn useful representations, recognize identities, and adapt to different environments. As the field progressed, researchers developed a variety of deep learning-based methods to address common challenges such as pose changes, lighting variation, occlusion, and lack of labeled data.

To better organize these developments, we group deep ReID methods into five main categories:  
(1) convolutional neural network (CNN)-based models,  
(2) generative adversarial network (GAN)-based models,  
(3) deep reinforcement learning (DRL)-based methods,  
(4) transformer-based models, and  
(5) large language model (LLM)-integrated methods.

Each category represents an important step in the development of ReID research. Newer methods were introduced to solve the limitations of earlier ones. For example, GANs were developed to handle variations in appearance across camera views, DRL offered adaptive learning strategies, and transformers helped capture complex visual relationships. Most recently, LLMs have been used to bring in textual understanding, allowing models to better recognize people through natural language descriptions. The following sections describe each category in detail, highlighting their contributions, challenges, and how they led to the next stage in this evolving field.

\subsubsection{CNN-based Siamese Architectures ReID}
\label{cnn}
CNNs marked the first deep learning wave in person ReID. These models replaced handcrafted descriptors with learnable hierarchical representations, enabling end-to-end optimization of feature extraction and identity matching.  Early CNN-based methods adopted Siamese~\cite{DYi2014,yxu2014} or triplet network structures, comparing image pairs or triplets to learn distance metrics for identity discrimination. Extensions like the cross-input neighborhood difference network~\cite{Ahmed} and LSTM-enhanced Siamese structures~\cite{varior_eccv_2} improved spatial feature alignment across views. Despite these gains, such architectures struggled with misalignments and lacked robustness to occlusions and scale variation. To address spatial misalignment and background noise, several attention-enhanced models emerged. \textit{HydraPlus-Net}~\cite{harmon_atten} introduced multi-directional attention, while \textit{MSCAN}~\cite{chen_multiscale_iccv} captured features at multiple receptive fields. Part-based approaches such as \textit{Part-based Convolutional Baseline + Refined Part Pooling (PCB+RPP)}~\cite{pcb} and \textit{Horizontal Pyramid Matching (HPM)}~\cite{hor_pyr_aaai2019} tackled misalignment by refining spatial partitioning, whereas the \textit{Omni-Scale Network (OSNET)}~\cite{osnet} introduced omni-scale feature learning to address scale variations. More sophisticated attention mechanisms emerged via \textit{Channel Attention Mechanism (CAM)}~\cite{bhuiyan2020pose}, \textit{Harmonious Attention Convolutional Neural Network (HA-CNN)}~\cite{liu_atten_arxiv2016}, \textit{Attentive but Diverse Network (ABD-Net)}~\cite{abdnet}, and \textit{Attention-Aggregation Network (AANet)}~\cite{aanet}, which improved resilience to occlusions and pose fluctuations. Semantic alignment further aided feature discrimination, as seen in \textit{SpindleNet}~\cite{spindlenet} and the \textit{Global-Local Alignment Descriptor (GLAD)}~\cite{wei_glad}, while \textit{MuDeep}~\cite{qian_tpami2019} employed a multi-scale deep architecture with leader-based attention layers to highlight significant regions. Additional progress came from \textit{Relation-Aware Global Attention (RGA)}~\cite{relation_aware}, which exploits pairwise correlations and affinities to capture global structure, and the \textit{Semantics Aligning Network (SAN)}~\cite{seman_jinaaai2020}, which fuses ReID and texture generation for enriched features. Finally, the \textit{Heterogeneous Local Graph Attention Network (HLGAT)}~\cite{heter_local} integrated graph-based learning to capture both intra- and inter-local relationships, further bolstering identity-preserving features across diverse conditions. CNN-based approaches laid a strong baseline but remained limited in modeling long-range dependencies and contextual semantics. These gaps motivated the shift toward generative and attention-based paradigms.

\subsubsection{GAN-based Approaches for ReID}
\label{gan}

GANs, first introduced by \cite{goodfellow} in 2014, have experienced rapid development and diversification with numerous variants and applications. GANs have played a crucial role in addressing style variance, pose variations, and resolution discrepancies in person ReID. Early work by \cite{vitro_gan_zheng2017} introduced GANs for generating unlabeled samples, which led to the \textit{Large-Scale Regularized Optimization (LSRO)}~\cite{huang_regu_tim2019} strategy to enhance model training with synthetic data. To tackle domain gaps, \textit{Person Transfer GAN (PT-GAN)}~\cite{person_transfer_weicvpr2018} and \textit{Similarity Preserving GAN (SPGAN)}~\cite{image_image_dengcvpr2018} adapted pedestrian images while maintaining identity consistency. For pose alignment, \textit{Pose Normalization GAN (PN-GAN)}~\cite{pose_qiancvpr2018} integrated pose structures with appearance features, improving generalization. \textit{Feature Distilling GAN (FD-GAN)}~\cite{fdgan_gearxiv} enhanced identity-related features while suppressing pose-dependent noise, whereas \textit{Multi-Pseudo Regularized Label (MpRL)}~\cite{huang_regu_tim2019} improved generalization by linking real and generated samples through virtual labels. \textit{Deformable GAN-based Network (DG-Net)}~\cite{joint_chencvpr2021} simultaneously learned identity-preserving feature transformations and generated realistic pedestrian images. Finally, \textit{DE+Uni}~\cite{fine_fanacmmm2018} optimized feature alignment by unifying real and GAN-generated samples for robust ReID. These advancements have significantly enhanced the adaptability of ReID models, making them more resilient to real-world variations.

 Despite their advantages, GAN-based ReID models face several limitations. Identity drift often occurs when generated images fail to preserve identity-specific features~, while mode collapse limits data diversity. Training instability and sensitivity to hyperparameters make results hard to reproduce. Generated samples also lack fine-grained semantic consistency~\cite{fdgan_gearxiv}, and performance gains are often dataset-specific, with poor generalization. These issues have encouraged the adoption of more stable alternatives such as Transformer and vision-language models.

\subsubsection{Deep Reinforcement Learning-based ReID }

DRL has emerged as a compelling approach for enhancing person ReID by enabling adaptive policy learning to address challenges such as viewpoint variations, occlusions, and noisy data~\cite{li2022overview}. Beyond the well-known \ textit{IDEAL (Identity Discriminative Attention Reinforcement Learning)}~\cite{lan2017deep}, which refines attention regions within bounding boxes to prioritize identity-discriminative features, several studies have expanded the DRL paradigm. Jiao and Bhanu~\cite{jiao2018deepagent} proposed DeepAgent, an integration framework where agents dynamically select preprocessing methods and feature extractors for each query image. Chen et al.\cite{chen2019self} introduced a self-critical attention mechanism guided by a critic module that refines attention maps using reinforcement signals, improving performance under weak supervision. Liu et al.\cite{liu2019deep} developed Deep Reinforcement Active Learning (DRAL), which selects high-value unlabeled samples for annotation using a reward strategy based on a sparse similarity graph, effectively reducing manual labeling efforts. In terms of feature selection, Shi et al.\cite{shi2020learning} proposed a refined attribute-aligned network (RAN) where agents choose optimal attribute features from distinct body parts to overcome spatial misalignments, while Zhang et al.\cite{zhang2020person} employed a reinforcement-guided attribute attention block (AAB) to discard noisy attributes and emphasize salient cues.

Despite these advances, DRL-based ReID models often suffer from unstable convergence, sensitivity to reward design, and scalability limitations when applied to large-scale or cross-domain scenarios. Moreover, DRL is typically integrated with discriminative models rather than functioning as a standalone solution, highlighting the need for more robust, interpretable, and scalable DRL frameworks tailored for real-world ReID applications.

\subsubsection{Transformer-based ReID} 
\label{transf}
Transformer-based \cite{vaswani2017attention} techniques have significantly advanced ReID approaches by capturing complex visual patterns and modeling long-range dependencies across different camera views. Recent progress in this area can be grouped into two main categories: Vision Transformer (ViT)-based models \cite{dosovitskiy2020image} and Swin Transformer-based models \cite{liu2021swin} (e.g., SOLDIER~\cite{chen2023beyond}). ViT-based approaches exploit global self-attention mechanisms to extract discriminative features, whereas Swin Transformer-based methods employ hierarchical window-based attention to efficiently capture both local and global information. 

\textbf{ViT-based Models.} A variety of ViT-based methods have emerged to strengthen representation learning. For instance, \textit{TransReID}~\cite{she2021} integrates a Jigsaw Patch Module (JPM) and Side Information Embeddings (SIE) to enrich feature diversity and incorporate camera viewpoint information. Building on this, \textit{DenseFormer}~\cite{dense_msiet22} directly connects all transformer layers to preserve finer details in feature flow, while \textit{ResT-ReID}~\cite{rest_chenpr22} combines ResNet-50 with transformer blocks and attention-guided Graph Convolution Networks (GCNs) to balance computation and depth. For occluded ReID, \textit{DRL-Net}~\cite{distan_jiaTM22} employs a global reasoning mechanism to identify individuals without explicit alignment. In contrast, \textit{TMGF (Transformer-based Multi-Grained Features)}~\cite{multi_liwacv23} utilizes contrastive learning for part-level and global representation, excelling in unsupervised settings. Addressing high-frequency texture details, \textit{TransReID+PHA}~\cite{pha_zhangcvpr23} integrates Patch-Wise High-Frequency Augmentation to refine fine-grained features. Meanwhile, \textit{DAAT (Dual-branch Adaptive Attention Transformer)}~\cite{daat_luivc23} employs selective token attention with Circle Loss for robust occlusion-aware learning, \textit{DCAL (Dual Cross-Attention Learning)}~\cite{dual_zhucvpr22} combines global-local and pairwise attention, and \textit{ABD-Net+Transformer} broadens attention mechanisms with adversarial learning to mitigate domain-specific biases.

\textbf{Improved Pretraining Strategies.} While many ViT-based architectures rely on ImageNet-pretrained backbones~\cite{Krizhevsky}, recent work has leveraged dedicated person-centric datasets for pretraining. \textit{PersonMAE}~\cite{hu2024personmae} presents a masked autoencoder approach trained on the large-scale LUPerson dataset \cite{fu2021unsupervised}, using cross-region multi-level prediction objectives to reach state-of-the-art performance in both supervised and unsupervised ReID tasks. Following a similar pipeline, \textit{PASS (Part-Aware Self-Supervised Pre-Training)}~\cite{zhu2022pass} employs learnable [PART] tokens and builds on the \textit{Distillation with no label (DINO)} framework~\cite{caron2021emerging}, emphasizing the distillation of local and global features for improved part-based learning. Extending this concept, \textit{PersonViT}~\cite{hu2025personvit} integrates Masked Image Modeling (MIM) into DINO to further refine local and global representations specifically designed for person ReID that achieves state-of-the-art ReID performance without requiring large labeled datasets, making it a highly scalable and generalizable self-supervised approach for person ReID.

\textbf{Swin Transformer-based Approaches.} In contrast, Swin Transformers adopt a hierarchical window-based attention scheme that efficiently models both local and global contexts, proving effective for real-world ReID applications. A notable example is  \textit{Semantic cOntrollable
seLf-supervIseD lEaRning (SOLIDER)}~\cite{chen2023beyond}, a self-supervised learning framework that integrates a semantic controller—via token clustering and semantic supervision—into a Swin Transformer backbone. This design adaptively balances appearance and semantic features, leading to robust representations for human-centric tasks.

Overall, both ViT-based and Swin Transformer-based paradigms continue to advance the field of person ReID, enhancing feature extraction, addressing occlusions, and improving cross-domain adaptability for practical applications in surveillance, security, and smart city infrastructures. Although Transformers significantly advanced the field by addressing occlusion, scale, and global reasoning, they still lacked semantic grounding in descriptive attributes, motivating the next transition to LLM-integrated ReID models.

\begin{table*}[t!]
    \centering
    \caption{Methods, References, and Performance Metrics on Market-1501 and DukeMTMC-ReID.}
\resizebox{0.7\textwidth}{!}{
    \begin{tabular}{l|c|l|c|c|c}
    \hline
    \textbf{ReID Methods} & \textbf{Reference} & \multicolumn{2}{c|}{\textbf{Market-1501}} & \multicolumn{2}{c|}{\textbf{DukeMTMC-reID}} \\ 
    \cline{3-6}
    &  & \textbf{mAP} & \textbf{R-1} & \textbf{mAP} & \textbf{R-1} \\ \hline
    
    \multicolumn{6}{l}{\textbf{CNN-Based Methods}} \\ \hline
    HydraPlus~\cite{hydraplus-net} & ICCV'17 & - & 76.9 & - & - \\
    MSCAN~\cite{li_latent_cvpr2017} & CVPR'17 & 66.7 & 86.8 & - & - \\

    DPFL~\cite{chen_multiscale_iccv} & ICCV'17 & 72.6 & 88.6 & 60.6 & 79.2 \\

   Gated Fusion~\cite{bhuiyan2020pose} & WACV'20 & 75.6 & 88.6 & 62.5 & 78.8 \\

    PCB + RPP~\cite{pcb} & ECCV'18 & 81.6 & 93.8 & 69.2 & 83.3 \\
    HPM~\cite{hor_pyr_aaai2019} & AAAI'19 & 82.7 & 94.2 & 74.3 & 86.6 \\
    OSNet~\cite{osnet} & TPAMI'22 & 84.9 & 94.8 & 73.5 & 88.6 \\
    CAM~\cite{yang_aug_cvpr2019} & CVPR'19 & 84.5 & 94.7 & 72.9 & 85.8 \\
    HA-CNN~\cite{harmon_atten} & CVPR'18 & 75.7 & 91.2 & 63.8 & 80.5 \\
    ABD-net~\cite{abdnet} & CVPR'19 & 85.0 & 95.1 & 78.6 & 89.0 \\
    AANet~\cite{aanet} & CVPR'19 & 83.4 & 93.9 & 79.0 & 87.7 \\
    Spindle~\cite{spindlenet} & CVPR'17 & 83.4 & 93.7 & 73.3 & 85.9 \\
    GLAD~\cite{wei_glad} & ACMMM'17 & 83.4 & 92.8 & 71.3 & 81.0 \\
    MuDeep~\cite{qian_tpami2019} & TPAMI'19 & 84.7 & 95.3 & 75.6 & 88.6 \\
    Pyramid~\cite{zheng_pyramid_cvpr_2019} & CVPR'19 & 88.5 & 95.7 & - & - \\
    RGA~\cite{relation_aware} & CVPR'20 & 84.2 & 96.1 & 79.5 & 91.0 \\
    SAN~\cite{seman_jinaaai2020} & AAAI'20 & 88.0 & 96.1 & 75.7 & 87.9 \\

    HLGAT~\cite{heter_local} & CVPR'21 & \textbf{93.4} & \textbf{97.5}  & \textbf{87.3} & \textbf{92.7} \\
    \hline

    \multicolumn{6}{l}{\textbf{GAN-Based Methods}} \\ \hline
    LSRO~\cite{vitro_gan_zheng2017} & ICCV'17  & 66.1 & 84.0 & 47.1 & 67.7 \\
    PT~\cite{liu_cvpr2018} & CVPR'18 & 68.9 & 87.6 & 56.9 & 78.5 \\
    PNGAN~\cite{pose_qiancvpr2018} & ECCV'18 & 72.6 & 89.4 & 53.2 & 73.6 \\
    FD-GAN~\cite{fdgan_gearxiv} & NIPS'18 & 77.7 & 90.5 & 64.5 & 80.0 \\
    MpRL~\cite{huang_regu_tim2019} & TIP'19 & 67.5 & 85.8 & 58.6 & 78.8 \\
    DG-Net~\cite{joint_zhang_cvpr2019} & CVPR'19 & 86.0 & 94.8 & 74.8 & 86.6 \\
    IDE-UnityStyle~\cite{style_liucvpr2020} & CVPR'20 & 89.3 & 93.2 & 65.2 & 82.1 \\ 
    \hline

    \multicolumn{6}{l}{\textbf{DRL-Based Methods}} \\ \hline

    IDEAL~\cite{lan2017deep} & BMVC'17  & 67.1 & 86.7 & - & - \\ 
DeepAgent~\cite{jiao2018deepagent} & ICIP'18 & - & 91.1 & - & - \\ 
DRAL~\cite{liu2019deep} & ICCV'19 & 66.3 & 84.2 & 56.0 & 74.3 \\ 
Self-Critical Attention~\cite{chen2019self} & ICCV'19 & 89.3 & 95.8 & 79.6 & 89.0 \\ 
RAN~\cite{shi2020learning} & Neurocomputing'20 & 89.2 & 95.6 & 79.1 & 88.9 \\ 
AAB~\cite{zhang2020person} & TIP'20 & 88.6 & 96.1 & 80.4 & 89.9 \\ \hline
    
        \multicolumn{6}{l}{\textbf{Transformer-Based Methods}} \\ \hline
    TransReID~\cite{she2021} & ICCV'21 & 88.9 & 95.2 & 82.0 & 90.7 \\
    Denseformer~\cite{dense_msiet22} & IET'22 & 88.1 & 95.0 & 81.3 & 90.0 \\
    ResT-ReID~\cite{rest_chenpr22} & PR'22 & 88.2 & 95.3 & 80.6 & 90.0 \\

    DAAT~\cite{daat_luivc23} & IVC'23 & 88.8 & 95.1 & 82.0 & 90.6 \\
    ABDNet+NFormer~\cite{nformer_wangcvpr22} & CVPR'22 & 93.0 & 95.7 & - & - \\ 
    DCAL~\cite{dual_zhucvpr22} & CVPR'22 & 87.5 & 94.7  & - & - \\

     PHA~\cite{pha_zhangcvpr23} & CVPR'23 & 90.2 & 96.1 & - & - \\
            TMGF Transformer~\cite{multi_liwacv23} & WACV'23 & 91.9 & 96.3 & 83.1 & 92.3 \\
        PASS~\cite{zhu2022pass} & ECCV'22 & 93.0 & 96.8 & - & - \\
         SOLDIER (Swin-B) ~\cite{chen2023beyond} & CVPR'23 & 93.6 & 96.9 & - & - \\ 
        PersonMAE~\cite{hu2024personmae} & TMM'25 & 93.6 & 97.1 & - & - \\
         PersonViT~\cite{hu2025personvit} & MVA,25 & \textbf{95.0} & \textbf{97.6} & \textbf{88.1} & \textbf{93.8} \\

    \hline

    \multicolumn{6}{l}{\textbf{LLM-Based Methods}} \\ \hline
    CLIP-ReID~\cite{lsi2023} & AAAI'23 & 90.5 & 95.4 & 83.1 & 90.8 \\
LVLM-ReID~\cite{wang2024large} & arXiv'24 & 89.2 & 95.6 & 82.8 & 92.2 \\
    
    MLLMReID~\cite{MLLMReID} & arXiv'24 & 91.5 & 96.5 & - & - \\
    PromptSG~\cite{yang2024pedestrian} & CVPR'24 & 94.6 & 97.0 & 81.6 & 90.0 \\
    MP-ReID~\cite{zhai2023multiprompts} & AAAI'24 & \textbf{95.5} & \textbf{97.7} & \textbf{88.9} & \textbf{95.7} \\ \hline
    \end{tabular}}
    \vspace{-3mm}
    \label{tab:deep}
\end{table*}

\subsubsection{Insights from Deep ReID before LLMs}

Early person ReID research focused on CNN-based models, evolving from simple Siamese architectures to more advanced networks that capture multi-scale and part-based features. For instance, \textit{PCB + RPP}~\cite{pcb} and \textit{OSNet}~\cite{osnet} use multi-branch or lightweight designs to improve accuracy. \textit{HLGAT}~\cite{heter_local} achieved strong results by modeling interactions between local regions using hierarchical graph networks. As shown in Table~\ref{tab:deep}, it reports 93.4\% mAP and 97.5\% Rank-1 on Market-1501, and 87.3\% mAP and 92.7\% Rank-1 on DukeMTMC-reID, demonstrating the high capacity of CNNs when enhanced with local graph modeling.

Meanwhile, GAN-based methods emerged to mitigate style variations and domain shifts through synthetic data generation. \textit{FD-GAN}~\cite{fdgan_gearxiv} and \textit{DG-Net}~\cite{joint_zhang_cvpr2019} generate realistic pedestrian images under varying conditions, while \textit{IDE-UnityStyle}~\cite{style_liucvpr2020} uses style-transfer to refine features. As summarized in Table~\ref{tab:deep}, IDE-UnityStyle achieves 89.3\% mAP and 93.2\% Rank-1 on Market-1501. However, despite strong individual results, GAN-based methods often exhibit inconsistent cross-domain generalization, highlighting the need for more robust feature learning.

Deep reinforcement learning (DRL)-based methods integrated sequential decision-making into ReID by modeling tasks such as attention policy learning, attribute selection, and noise filtering. Approaches like \textit{IDEAL}~\cite{lan2017deep}, \textit{DRAL}~\cite{liu2019deep}, and \textit{AAB}~\cite{zhang2020person} exemplify this paradigm. Table~\ref{tab:deep} shows that AAB reaches 88.6\% mAP and 96.1\% Rank-1 on Market-1501, a strong performance among DRL approaches. Nevertheless, these methods are limited by reward design complexity, training instability, and poor interpretability—issues that hinder broader deployment and integration into language-grounded systems.

Transformer-based methods have recently outperformed previous paradigms by leveraging global attention to capture fine-grained spatial and semantic dependencies. For instance, \textit{TMGF Transformer}~\cite{multi_liwacv23} achieves 91.9\% mAP and 96.3\% Rank-1 on Market-1501, as listed in Table~\ref{tab:deep}. Recent models such as \textit{PASS}~\cite{zhu2022pass}, \textit{SOLDIER}~\cite{chen2023beyond}, and \textit{PersonMAE}~\cite{hu2024personmae} push the performance further. Notably, \textit{PersonViT}~\cite{hu2025personvit} achieves 95.0\% mAP and 97.6\% Rank-1 on Market-1501, and 88.1\% mAP and 93.8\% Rank-1 on DukeMTMC-reID—representing the current state-of-the-art (Table~\ref{tab:deep}). These results reinforce the effectiveness of vision Transformers in encoding identity-relevant features across varying scenes and viewpoints.

\subsubsection{Limitations of pre-LLM Deep ReID and the Motivation for LLM Integration}

Despite the remarkable progress across CNN-based, GAN-based, DRL-based, and Transformer-based ReID models, a common limitation persists: most approaches still rely heavily on coarse-grained visual features or manually defined semantic attributes, often overlooking the rich contextual information that could help distinguish visually similar individuals. For instance, while these models may accurately capture appearance-related traits such as clothing color or pose, they often fall short in utilizing nuanced descriptions like “a person wearing a faded denim jacket with a visible tear on the sleeve.” Additionally, while DRL-based methods introduce adaptability, they typically lack interpretability and semantic reasoning, making it difficult to understand or justify model decisions in human terms.

This limitation highlights the need for more expressive and semantically rich frameworks, paving the way for the integration of LLMs. LLMs offer the ability to generate and refine fine-grained textual descriptions, align vision-language representations, and support natural-language interactions. By bridging the gap between raw visual input and high-level semantic understanding, LLMs enable more robust, generalizable, and explainable ReID pipelines. The following sections explore how LLM-based approaches leverage these advantages to advance the state-of-the-art in person ReID.

\subsubsection{Integration of LLMs in ReID}
\label{LLM}

LLMs have gained significant attention for their ability to handle a wide range of Natural Language Processing (NLP) tasks. Their skill in processing textual prompts has motivated researchers to combine them with large vision models, creating powerful systems for vision-language tasks. This idea started with CLIP~\cite{clip}, which used contrastive learning on large-scale image-text pairs to align visual and textual representations. Building on CLIP, multimodal LLMs~\cite{zhu2023minigpt,liu2024visual,dai2024instructblip,bai2023qwen,chen2023sharegpt4v} have been developed, improving vision-language capabilities by training on extensive image-text datasets and refining prompt engineering methods. 

Recent research has explored using these multimodal models for person ReID, where combining textual descriptions with visual features through a two-stage training process has improved identity matching~\cite{lsi2023}. However, challenges such as overfitting to specific prompts~\cite{yang2024mllmreid} and the limited availability of labeled datasets~\cite{chen2023unveiling} have led researchers to explore new strategies, including cluster-aware prompt learning and unsupervised learning techniques to improve generalization. More recently, \cite{yang2024pedestrian} showed that CLIP-ReID models mainly rely on predefined soft prompts, making them less effective for recognizing unseen identities. To overcome this, the \textit{Prompt-driven Semantic Guidance (PromptSG)} framework was introduced, using an inversion network to generate pseudo tokens that capture visual context and create personalized language descriptions, enhancing model adaptability. The effectiveness of prompt learning in ReID has also encouraged the use of multiple prompts to capture fine-grained visual details~\cite{zhai2023multiprompts}. For example, \textit{Multi-Prompt ReID (MP-ReID)} \cite{zhai2023multiprompts} combined explicit prompts generated by LLMs like ChatGPT with implicit learnable prompts, using cross-modal alignment techniques to better link visual and textual features, achieving state-of-the-art results in ReID benchmarks. 

Additionally, datasets such as \textit{Instruct-ReID}~\cite{he2023instructreid}, its extended version \textit{Instruct-ReID++}~\cite{he2024instructreid}, and \textit{MLLMReID}~\cite{MLLMReID} have been introduced to evaluate ReID in multi-task learning settings. A recent framework, LVLM-ReID~\cite{wang2024large}, utilizes \textit{Large Vision-Language Models (LVLMs)} to generate a single semantic token that encodes key pedestrian attributes and refines it through a semantic-guided interaction module. Instead of relying solely on traditional ReID metrics like Rank-1 accuracy and mean Average Precision (mAP), Hambarde et al.~\cite{hambarde2025human} evaluated leading LVLMs - such as GPT-4o~\cite{gpt4_technicalreport} and  Gemini 2.0 \cite{team2023gemini} - on Human ReID tasks. Their study introduced a structured evaluation pipeline that includes dataset curation, prompt engineering, and robust evaluation metrics such as similarity scores, precision, recall, F1 score, and area under the curve (AUC).
While fine-tuning visual encoders can enhance supervised performance, it often reduces zero-shot generalization. This highlights the need to further explore the zero-shot capabilities of multimodal LLMs for ReID~\cite{yang2024pedestrian}.

Furthermore, cross-modal ReID tasks, particularly text-to-image ReID, have utilized LLMs to address issues of noisy descriptions in benchmark datasets~\cite{tan2024harnessing,li2024data}. In~\cite{tan2024harnessing}, LLMs generate diverse and contextually rich textual descriptions, mitigating text noise and model overfitting by using multi-turn dialogues and template-based text generation. Similarly, Li et al.~\cite{li2024data} introduced an LLM-based data augmentation method that rewrites textual descriptions to increase vocabulary diversity and improve sentence structure while preserving semantic consistency. Although cross-modal ReID is an important direction, this survey focuses primarily on single-modality ReID, with cross-modal approaches briefly discussed in relevant sections.

Among Transformer-based approaches, as highlighted in Table~\ref{tab:deep}, \textit{PersonViT}~\cite{hu2025personvit} achieves a 95.0\% mAP and 97.6\% Rank-1 on Market-1501, whereas the LLM-based \textit{MP-ReID}~\cite{zhai2023multiprompts} slightly exceeds it with 95.5\% mAP and 97.7\% Rank-1. A similar trend emerges on DukeMTMC-reID, indicating that LLMs can leverage textual cues to capture subtle appearance details beyond what purely visual models can detect. This additional linguistic context leads to stronger adaptability and improved accuracy across diverse scenarios.

In summary, we can deduce that prompt-based interaction serves as the core mechanism that enables LLMs to adapt flexibly to varying ReID scenarios. Unlike conventional models that rely solely on fixed visual embeddings, prompt-driven methods allow the ReID pipeline to guide attention toward identity-relevant cues-such as clothing texture, accessories, or behavioral context-through natural language. This semantic guidance not only improves interpretability but also empowers models to generalize across datasets with minimal supervision. As a result, leveraging prompts has become a powerful strategy for bridging the gap between low-level appearance features and high-level identity semantics in person ReID.

\section{Leveraging Extended Prompts in the CLIP-ReID Framework}
\label{extend_prompt}

\begin{figure*}[t!]
\centering
\includegraphics[width=0.9\linewidth,,trim={0in 0in 0in 0in},clip]{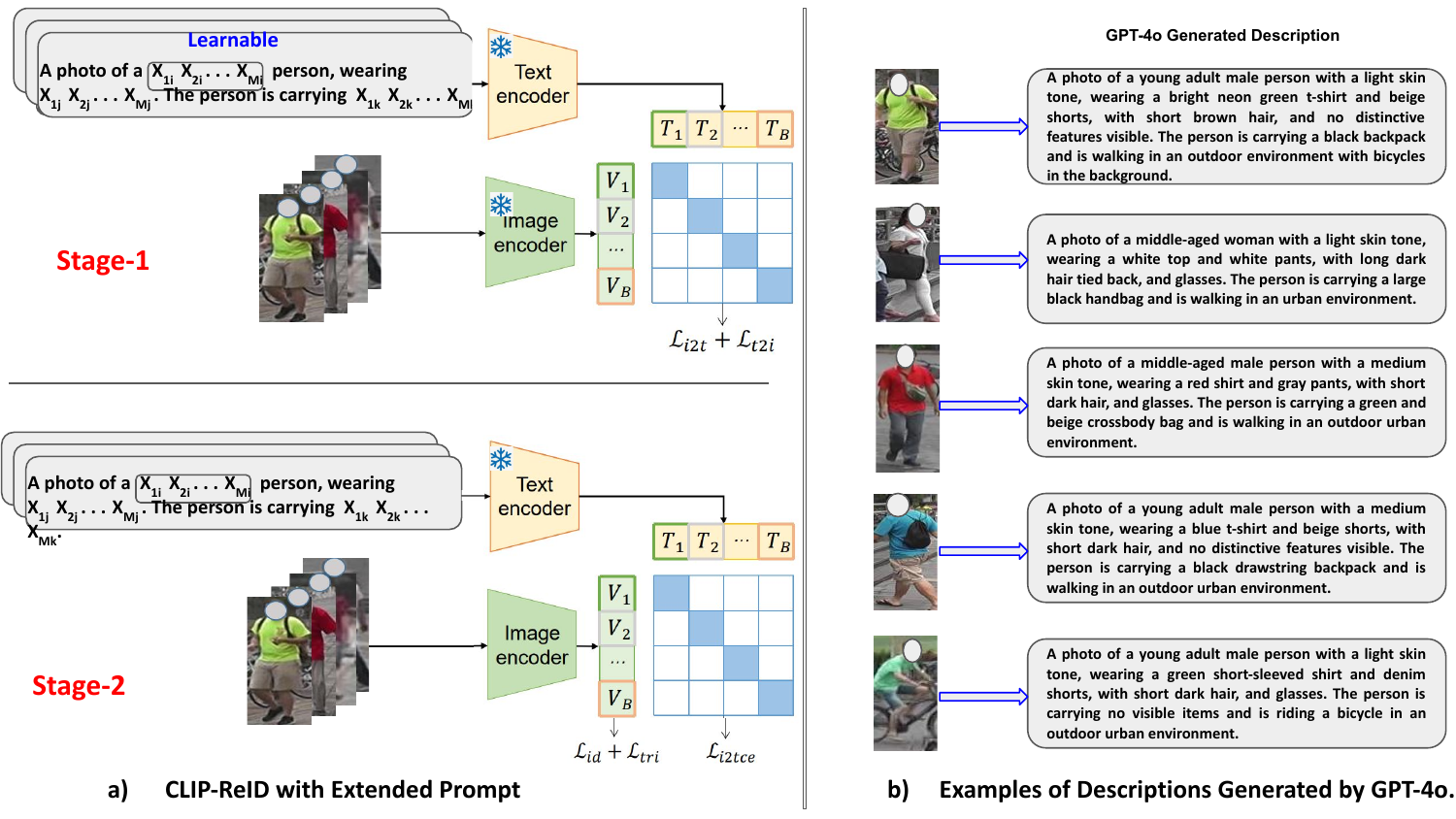}

   \caption{ a) Illustrating CLIP-ReID~\cite{lsi2023} with Extended Prompts (Stages 1 and 2); b) Examples of Descriptions Generated by GPT-4o.
   }
\label{fig:extended_prompts}

\end{figure*}

In addition to conducting a survey, we extend the standard CLIP-ReID framework to investigate how variations in {prompts} affect the results. For this purpose, we introduce \textit{extended dynamic prompts}. In our proposed extended dynamic prompting technique, we leverage GPT-4o \cite{gpt4_technicalreport}, to generate a description of each image in the dataset. In this way, we generate a detailed description of the given image which is not static, but rather dynamic in nature. These prompts are specifically designed to enhance the model's capacity to connect visual and textual data, thereby providing richer and more contextually accurate descriptions for each identity. As illustrated in Figure~\ref{fig:extended_prompts}, the introduction of extended prompts significantly boosts the model's discriminative ability, enabling it to more effectively distinguish between different identities.

\subsection{Stage 1: Learning ID-Specific Tokens with Extended Prompts}

The first stage of our framework mirrors the approach of the standard CLIP-ReID~\cite{lsi2023} but with a crucial enhancement: the integration of extended learnable tokens. These tokens are introduced to address the inherent ambiguities present in textual descriptions of each identity. Unlike traditional CLIP-ReID models, where text descriptions are static and predefined, our method dynamically optimizes these tokens to align closely with the unique characteristics of each identity.

In this stage, we initialize a set of learnable tokens, denoted as \([X]_m\), where \(m \in \{1, 2, \dots, M\}\). These tokens represent various aspects of the identity, such as appearance, clothing, and accessories. They are embedded within a fixed textual template, expressed as:

\emph{``A photo of a \([X]_{1i} [X]_{2i} \dots [X]_{Mi}\) person, wearing \([X]_{1j} [X]_{2j} \dots [X]_{Mj}\). The person is carrying \([X]_{1k} [X]_{2k} \dots [X]_{Mk}\)”.}

Here, the placeholders \([X]_m\) correspond to learnable tokens that encapsulate specific features like the color of clothing, types of accessories, or even the surrounding environment. The integration of these details allows the model to create a more comprehensive and nuanced embedding for each identity. These placeholders are optimized during training, refining their embeddings to capture the distinctive attributes of the person being described.

The contrastive loss functions \(\mathcal{L}_{i2t}\) and \(\mathcal{L}_{t2i}\) are employed to align the embeddings of images and text. Specifically, \(\mathcal{L}_{i2t}\) ensures that the image embedding closely matches its corresponding text embedding, while \(\mathcal{L}_{t2i}\) enforces the reverse alignment. Following the same training strategy as in CLIP-ReID~\cite{lsi2023}, this stage primarily focuses on optimizing the learnable tokens based on  \(\mathcal{L}_{i2t}\) and \(\mathcal{L}_{t2i}\) while keeping the CLIP model parameters fixed, enhancing the model’s ability to effectively leverage the textual modality.

\subsection{Stage 2: Enhancing Model Capability through Extended Descriptions}

After completing the first stage, the model has learned ID-specific tokens that capture essential identity features, such as clothing color, types of accessories, and surrounding context. In the second stage of training, the prompt learner and the text encoder are frozen, and only the parameters of the image encoder within the CLIP model are further optimized. This optimization uses a combination of loss functions: identity loss \(\mathcal{L}_{id}\), triplet loss \(\mathcal{L}_{tri}\), and cross-entropy loss \(\mathcal{L}_{i2tce}\). 

The identity loss \(\mathcal{L}_{id}\) is employed to maximize the likelihood of accurately predicting the correct identity class. Concurrently, the triplet loss \(\mathcal{L}_{tri}\) ensures that the distance between embeddings of positive pairs (same identity) is smaller than the distance between negative pairs (different identities) by a margin \(\alpha\). Additionally, the cross-entropy loss \(\mathcal{L}_{i2tce}\) further refines the alignment between image and text embeddings, leveraging the text features obtained during the first stage.

\subsection{Comparison with GPT-4o Generated Descriptions}

Beyond manually crafted extended prompts, we also explore the use of descriptions generated by LLMs, such as GPT-4o \cite{gpt4_technicalreport}, as an alternative. These LLM-generated descriptions are created based on the visual content of each image, as depicted in Figure~\ref{fig:extended_prompts}b. While these generated descriptions can provide detailed and contextually rich narratives, they often lack the specificity and controllability that our manually designed \textit{extended prompts} offer. \\

The GPT-4o descriptions are generated by feeding the image into a pre-trained language model, which then produces a narrative based on the visual features detected. Although this method can yield descriptive and nuanced text, it may introduce variability and noise, which could degrade the performance of the ReID system. In contrast, our \textit{extended prompts} are meticulously designed to focus on key identity-specific features, which are essential for maintaining high accuracy in ReID tasks. Within the CLIP-ReID model, these descriptions are utilized as prompts, reinforcing the model's capability to associate images with precise textual descriptions.

\subsection{Comparative Analysis of extended Prompt Variations in CLIP-ReID with ViT-16}

\noindent For this experimental analysis, we adopt the ViT-B/16 model from CLIP with a patch stride of 8 to improve spatial resolution. The transformer has 12 layers with a hidden size of 768, and its output is projected to 512 dimensions via a linear layer to match the text encoder. In the first stage, only the learnable text tokens are optimized using the Adam optimizer with an initial learning rate of $3.5 \times 10^{-4}$, decayed by a cosine schedule, and a batch size of 64 is used without data augmentation. In the second stage, the image encoder is fine-tuned using the Adam optimizer. Each training batch includes $B = P \times K$ images, where $P = 16$ identities and $K = 4$ samples per identity. Standard data augmentation methods are applied, including random horizontal flipping, padding, cropping, and erasing. The learning rate is linearly increased from $5 \times 10^{-7}$ to $5 \times 10^{-6}$ over the first 10 warm-up epochs, then decreased by a factor of 0.1 at the 30\textsuperscript{th} and 50\textsuperscript{th} epochs. The ViT model is trained for a total of 60 epochs.

\begin{table*}[t!]
\centering
\caption{Comparison of prompt variations in CLIP-ReID using ViT-16 as the backbone, evaluated on Market1501 and DukeMTMC datasets.}
\resizebox{1.0\textwidth}{!}{
\begin{tabular}{|c|c|c|c|c|c|c|}
\hline
\textbf{Methods}  & \textbf{Prompts} & \textbf{CTX} & \multicolumn{2}{c|}{\textbf{Market1501}} & \multicolumn{2}{c|}{\textbf{DukeMTMC}} \\ 
 &   &  & \textbf{Rank-01} & \textbf{mAP} & \textbf{Rank-01} & \textbf{mAP} \\ \hline
\multirow{10}{*}{CLIP-ReID} & `` " & 4 & 93.8 & 87.8 & 89.9 & 81.9 \\ \cline{2-7}
 \multirow{10}{*}{with ViT-16} & `` X X X X " & 4 & 94.4 & 88.8 & 90.8 & 82.4 \\ \cline{2-7}
   & ``A photo of a X X X X person" & 4 & 94.9 & 89.5 & 90.6 & 82.7 \\ \cline{2-7}
    & GPT-4o Generated Descriptions & 4 & 94.1 & 86.8 & 89.2 & 78.8 \\  \cline{2-7}
   & ``A photo of a X X X X person, wearing X X X X" & 4 & 94.7 & 89.4 & 90.3 & 82.5 \\ \cline{2-7}

    & ``A photo of a X X X X person" & 8 & 95.1 & 89.4 & 90.5 & 82.5 \\ \cline{2-7}

   & ``A photo of a X X X X person, wearing X X X X" & 8 & 95.2 & 89.7 & 90.8 & 82.7 \\ \cline{2-7}

    & ``A photo of a X X X X person, wearing X X X X. The person is carrying X X X X." & 8 & 95.1 & 89.6 & 90.6 & 82.7 \\ \cline{2-7}

     & ``A photo of a X X X X person, wearing X X X X. The person is carrying X X X X." & 10 & 95.3 & 89.6 & 90.8 & 82.5 \\ \cline{2-7}

    & ``A photo of a X X X X person, wearing X X X X. The person is carrying X X X X." & 20 & 95.1 & 89.5 & 91.1 & 82.8 \\  \cline{2-7}

      & ``A photo of a X X X X person." & 20 & 94.1 & 88.3 & 89.6 & 82.0 \\ \cline{2-7}
      
     & GPT-4o Generated Descriptions & 20 & 94.1 & 89.4 & 90.8 & 82.6 \\  \cline{2-7}

         & ``A photo of a X X X X person" & 35 & 93.9 & 88.1 & 89.2 & 81.7 \\ \cline{2-7}

          & GPT-4o Generated Descriptions & 35 & \textbf{95.7} & \textbf{90.4} & \textbf{91.3} & \textbf{83.6} \\ \hline
\end{tabular}}

\label{tab:clip_reid_prompts}
\vspace{-.4cm}
\end{table*}

Table~\ref{tab:clip_reid_prompts} shows a detailed comparison of prompt variations used in CLIP-ReID with ViT-B/16, highlighting the impact of prompt design and context length (CTX) on performance across Market1501 and DukeMTMC datasets. Starting with the baseline configuration that uses no text prompting (``\quad"), the model achieves 93.8\% Rank-1 and 87.8\% mAP on Market1501, and 89.9\% Rank-1 with 81.9\% mAP on DukeMTMC. Introducing a simple prompt like ``A photo of a X X X X person'' (CTX = 4) leads to a notable improvement of +1.1\% Rank-1 and +1.5\% mAP on Market1501, and +0.8\% Rank-1 and +0.5\% mAP on DukeMTMC. Adding clothing attributes (e.g., ``wearing X X X X'') and increasing CTX to 8 further enhances performance to 95.2\% Rank-1 and 89.4\% mAP on Market1501, marking an overall gain of +1.4\% Rank-1 and +1.6\% mAP from the baseline. DukeMTMC also benefits similarly with +0.8\% Rank-1 and +0.6\% mAP compared to the simple prompt. These results indicate that moderate prompt enrichment yields consistent performance gains, especially when clothing cues are included.

More complex prompts that add object-related details (e.g., ``The person is carrying X X X X'') show marginal benefits. For example, with CTX = 10, the Rank-1 on Market1501 is 95.0\%, only +0.2\% over the CTX = 8 setting. Interestingly, GPT-4o-generated descriptions at CTX = 20 do not outperform handcrafted prompts, yielding 94.8\% Rank-1 and 88.8\% mAP on Market1501, which is +1.0\% Rank-1 and +1.0\% mAP over the baseline but still slightly behind the simpler crafted prompts. However, when the context length is increased to 40, GPT-4o-generated prompts achieve the highest results overall, with 95.7\% Rank-1 and 90.4\% mAP on Market1501, and 91.3\% Rank-1 and 83.6\% mAP on DukeMTMC—reflecting an absolute improvement of +1.9\% Rank-1 and +2.6\% mAP on Market1501 over the baseline. This highlights that well-structured, semantically rich prompts—especially those generated by LLMs and paired with sufficient context—can significantly improve ReID performance. In contrast, simply increasing CTX without meaningful textual content (as seen in the baseline with CTX = 40) shows no improvement, confirming the need for dynamic and informative descriptions.

Although the improvement margin is small, this finding demonstrates a potential method to enhance ReID accuracy through the use of GPT-4o-generated descriptions. To support further research and exploration, we are making the GPT-4o-based person descriptions for both datasets publicly available. This approach not only highlights the benefits of incorporating advanced language models in prompt design but also opens up new avenues for improving ReID accuracy through carefully crafted descriptions.

\section{Current Challenges and Future Directions}
\label{Future}
This section outlines the main challenges in person ReID, including occlusions, domain adaptation, scalability, and reliance on high-quality textual descriptions in LLM-based models. It also highlights potential solutions and future research directions to improve ReID performance in real-world applications. A summary of these challenges, solution strategies, and potential future directions is provided in Table~\ref{chall}.

\subsection{Current Challenges}

\begin{itemize}
\item \textbf{Clothes-Variation Challenge in Lifelong Person ReID:} One of the major challenges in person ReID, especially in long-term or real-world surveillance, is handling changes in a person’s clothing. Most existing ReID methods work well only when a person wears the same outfit across different cameras. But in real life, people change clothes frequently, which makes it hard for these systems to recognize them. This problem has led to a new research area called clothes-changing ReID (CC-ReID)~\cite{wang2025content,siddiqui2024dlcr}. In standard settings, ReID models~\cite{wang2025content} can achieve very high accuracy-almost 99--100\% Rank-1, but when clothes change, accuracy drops sharply to around 64--66\%. The issue gets worse when facial features are blurred for privacy, removing another important cue for recognition. Also, most datasets used for CC-ReID have limited clothing variety, making it hard for models to learn how to handle diverse appearances. New methods like Content
and Salient Semantics Collaboration (CSSC) ~\cite{wang2025content} and Diffusion and LLMs for CCReID (DLCR)~\cite{siddiqui2024dlcr} try to solve these problems by learning clothing-invariant features or generating diverse training images with AI, but the task remains very challenging.

    \item \textbf{Handling Occlusion and Pose Variations:} Despite significant progress in the field of person ReID, occlusion and pose variability continue to present substantial challenges. A specialized research area known as Occluded ReID has emerged to specifically address these issues. According to a recent survey on Occluded ReID~\cite{ning_occ24}, the performance of state-of-the-art models on occluded ReID benchmarks is still limited, with Rank-01 accuracy ranging between 65\% and 70\%. In contrast, the performance on the same datasets without occlusion reaches a much higher Rank-01 accuracy, typically between 88\% and 95\%. These findings indicate that even the most advanced ReID models struggle to accurately identify individuals when they are partially obscured or when there are significant variations in pose between images. This highlights the ongoing need for further research and development to improve the robustness of ReID systems under these challenging conditions.

\begin{table*}[t!]
    \centering
    \caption{Summary of Current Challenges, Solutions, and Future Research Directions in ReID}
    \resizebox{1.0\textwidth}{!}{
    \begin{tabular}{|p{6cm}|p{6cm}|p{6cm}|}
    \hline
    \textbf{Challenges} & \textbf{Solution Strategies} & \textbf{Potential Future Directions} \\ \hline
    \textbf{Scalability} & Optimized model architectures~\cite{osnet,masson_bhuiyan2021,Remigereau_2022} and few-shot learning approaches to reduce data requirements. & Focus on model compression techniques (e.g., pruning, quantization) and further development of few-shot and zero-shot learning to reduce computational costs. \\ \hline
    \textbf{Generalization Across Datasets} & Domain adaptation~\cite{mekhazni_eccv2020,zhong_pami,Feng_TIP2021, Yu_pami2018, Liu_TCSVT2022,Li_cviu2023,chen_pr2023} and Generalization techniques~\cite{jiao_eccv2022, gong_tcvst2023,bhuiyan_cviu2024}, such as adversarial training to transfer learning between domains. & Research into unsupervised domain adaptation and continual learning to improve generalization across unseen domains without requiring extensive labeled data. \\ \hline
    \textbf{Occlusion and Pose Variation} & Integration of pose estimation models and partial ReID~\cite{sun_part_cvpr2019,ning_occ24} methods that handle occluded pedestrian images. & Development of advanced pose-invariant models and multimodal approaches combining visual and textual data to improve identification in complex scenarios. \\ \hline
    \textbf{High Computational Cost} & Use of CNN-based models~\cite{fastreid} and transformer-based approaches that are more efficient in feature extraction. & Explore lightweight architectures and energy-efficient hardware accelerators for real-time deployment. \\ \hline
    \textbf{Clothes-Changing ReID} & Learning identity-invariant features using semantic-aware models~\cite{wang2025content} or synthetic data from generative methods like diffusion models~\cite{siddiqui2024dlcr}. & Further integration of diffusion models to generate diverse, identity-preserving images, and improve training under appearance variation with limited supervision. \\ \hline
    \textbf{Quality of Textual Prompts in LLM-guided ReID} & Use of LLMs for generating textual descriptions and prompts~\cite{zhai2023multiprompts,he2023instructreid}. & Research into automated and dynamic prompt generation for LLMs to enhance the quality and context of descriptions in challenging environments. \\ \hline
    \end{tabular}}
    \label{chall}
    \vspace{-0.4cm}
\end{table*}

    \item \textbf{Domain Adaptation and Generalization:} One of the central challenges in person ReID is the performance drop observed when models trained on one dataset are tested on another, primarily due to domain shifts caused by variations in camera quality, lighting, and backgrounds. While many supervised ReID methods excel within their training domains, they often lack generalization to unseen settings. To mitigate this, two key subfields have emerged: domain adaptation ReID, which adapts models using unlabeled target-domain data~\cite{mekhazni_eccv2020,zhong_pami,Feng_TIP2021,Yu_pami2018,Liu_TCSVT2022,Li_cviu2023,chen_pr2023}, and domain-generalizable ReID, which aims to generalize without any target-domain data~\cite{jiao_eccv2022,gong_tcvst2023,bhuiyan_cviu2024}. Despite progress in unsupervised domain adaptation, domain-generalizable approaches still lag behind supervised counterparts, underscoring the need for robust ReID models that can generalize effectively across diverse environments.

\item \textbf{Reproducibility of the Exiting SOTA approach}
    One of the significant challenges highlighted in the literature is the limited availability of source code for state-of-the-art approaches such as MP-ReID~\cite{zhai2023multiprompts} and MLLMReID~\cite{MLLMReID}. Despite these methods achieving notable results, the lack of accessible code presents a barrier to reproducing their reported performance. In our efforts to validate these approaches, we implemented them independently based on the details provided in the respective papers. However, our reproduced results did not match the accuracy reported by the original authors. We also made attempts to contact the authors to request access to the source code, but, unfortunately, we have not yet received a response. This lack of reproducibility and transparency highlights a broader issue in the field and underscores the importance of making research code publicly available to ensure the reliability and advancement of future work.

       \item \textbf{Dependency on Quality of Textual Descriptions:} The performance of LLM-integrated ReID systems is highly dependent on the quality and contextual relevance of textual descriptions, which are critical for the accurate alignment of text images in multimodal processing. Poorly crafted prompts, whether ambiguous, generic, or noisy, can degrade performance and lead to misidentification, especially in complex environments. For example, as shown in Table~\ref{tab:clip_reid_prompts}, a vague prompt like “wearing a red shirt” lacks sufficient discriminatory power. This highlights the need for advanced techniques to generate, filter, or adapt prompts based on visual context, thereby improving the robustness and reliability of LLM-guided ReID in real world scenarios.
    
    \item \textbf{Scalability Issues:} Scalability is a major challenge for person ReID models, especially as datasets in surveillance systems grow larger and more complex. Current models are often focused on maximizing accuracy, but this comes at the cost of high computational requirements, making them difficult to scale in real-time applications. The need for significant processing power, memory, and energy limits their use in resource-constrained environments like edge computing. Real-time processing, crucial for surveillance, becomes even more challenging with these models. To address this, ReID models need to be optimized for efficiency without losing accuracy. This could involve new algorithms, model compression, parallel processing, hardware acceleration, and smarter data handling methods like feature selection. Overcoming these challenges is essential for deploying ReID systems in a wide range of real-world applications.
\end{itemize}

\subsection{Future Directions and Recommendations}

\begin{itemize}

\item \textbf{Diffusion Models for Clothes-Changing ReID:} Diffusion models offer a promising direction for improving clothes-changing person ReID. By generating diverse images of the same person in different outfits while keeping identity traits intact, they help models learn features that are not dependent on clothing. Recent work by Siddiqui et al.~\cite{siddiqui2024dlcr} shows that text-guided diffusion can create large, high-quality training sets that significantly boost recognition accuracy in challenging clothes-changing scenarios. Future research can build on this by combining generative data with identity-aware training to develop more robust and generalizable CC-ReID systems.

        \item \textbf{Enhanced Multimodal Integration:} Enhanced multimodal integration is crucial for advancing ReID systems, and future research should focus on developing more sophisticated methods for combining visual and textual data. This involves creating advanced LLMs specifically designed for ReID tasks, capable of generating detailed and contextually relevant descriptions that align closely with visual inputs. Such integration can significantly improve model accuracy, especially in challenging scenarios like occlusions or varying poses. Additionally, developing dynamic fusion techniques that adaptively integrate visual and textual information based on task context can further enhance performance, making ReID systems more accurate and reliable in real-world applications.

      \item \textbf{Zero-Shot Learning and Generalization with LLMs:} Exploring zero-shot learning in LLMs for ReID tasks is crucial for future research, as it allows models to identify individuals or objects they haven't seen before. This is especially important for real-world applications with high variability. To achieve strong zero-shot performance, better techniques for prompt generation are needed, which can capture the unique features of identities and contexts without prior training. This might involve creating adaptive, dynamic prompts or using more complex prompt structures. Pre-training on diverse datasets can also improve the model's ability to generalize. By tackling these challenges, LLM-based ReID systems can become more adaptable and effective in real-world situations.

    \item \textbf{Adaptive Learning Strategies for LLM Integration:} Future research should focus on developing adaptive learning strategies for LLMs to generate and refine prompts in real-time, improving ReID systems' adaptability to diverse scenarios. One approach is to incorporate human-in-the-loop methods, where the model learns from user feedback to refine its prompt generation over time. By leveraging real-time input, the model will be able to adapt to specific tasks, such as identifying individuals under changing conditions or interpreting ambiguous visual cues
   
    \item \textbf{Enhanced Occlusion Handling with Advanced Attention Mechanisms:} Future research could focus on developing more sophisticated attention mechanisms that dynamically adjust to images' occluded or partially visible regions. This could involve integrating attention models with deep graph structures to better model inter-local and intra-local relationships.

    \item \textbf{Scalable and Real-Time ReID Models:} Future research should focus on reducing the computational demands of ReID models to enhance their scalability and suitability for real-time applications, particularly in environments where quick processing and decision-making are critical. As ReID models become increasingly complex, integrating robust LLMs offers a powerful approach to improving accuracy and contextual understanding. However, the challenge lies in maintaining this sophistication while ensuring the models remain computationally efficient. One promising direction is to optimize model architectures through techniques like model pruning~\cite{masson_bhuiyan2021,prune_2,cheng_prun_pami} or knowledge distillation~\cite{Remigereau_2022, know_02,know_03,know_04}, where the knowledge from larger, more complex LLM-integrated models is transferred to smaller, more efficient versions. This would allow the development of lightweight ReID models that retain the robustness and accuracy of their larger counterparts but are better suited for real-time deployment in resource-constrained environments. By balancing computational efficiency with the advanced capabilities of LLM integration, these efforts could lead to ReID systems that are not only powerful but also practical for widespread, real-time use.
    
\end{itemize}

\section{Conclusions}
\label{con}
This paper presents the first comprehensive survey focused on the emerging role of large language models (LLMs) in person ReID, highlighting a shift from traditional visual-only techniques to multimodal approaches that incorporate language. While earlier surveys emphasized handcrafted features, convolutional networks, and domain-specific methods, our work captures the growing influence of language-driven systems. By tracing the historical development of ReID from basic descriptors to deep learning and now to vision language integration, we provide a structured overview of how the field has progressed and where it is heading. We also outline key research challenges such as effective prompt generation, multimodal domain adaptation, and vision-language alignment. As part of this effort, we generate and publicly share a large collection of natural language descriptions created using GPT-4o for two widely used ReID benchmarks. These image specific descriptions add semantic depth and support further research on language-guided ReID methods. Beyond the survey, we demonstrate how these extended prompts can be integrated into the CLIP-ReID pipeline to improve visual-textual association. Our results show that prompt-based methods enhance identity matching, especially in challenging scenarios involving occlusion, pose variation, or similar appearances. Together, this survey, the GPT-4o-based descriptions, and our experimental insights open new directions for building the next generation of ReID systems that are more robust, adaptable, and semantically aligned with human-level understanding.

\section*{ACKNOWLEDGMENTS}

This research is supported by research grants from the Natural Sciences and Engineering Research Council (NSERC) of Canada (RGPIN-2020-07157 and RGPIN-2019-06799), York Research Chairs (YRC) program, and Dapasoft Inc.  We also acknowledge the Digital Research Alliance of Canada for providing us with the computing resources to conduct experiments. We extend our sincere gratitude to the editors for their time, insightful feedback, and valuable suggestions, which have significantly enhanced the quality of this work.

{
    \small
    \bibliographystyle{ieeenat_fullname}
    \bibliography{main}
}


\end{document}